\documentclass[sigconf]{acmart}
\AtBeginDocument{%
  }

\copyrightyear{2026}
\acmYear{2026}
\setcopyright{cc}
\setcctype{by}
\acmConference[MM '26]{Proceedings of the 34th ACM International Conference on Multimedia}{November 10--14, 2026}{Rio de Janeiro, Brazil}
\acmBooktitle{Proceedings of the 34th ACM International Conference on Multimedia (MM '26), November 10--14, 2026, Rio de Janeiro, Brazil}
\acmDOI{10.1145/3767308.3836590}
\acmISBN{979-8-4007-2213-4/2026/11}

\usepackage{algorithm}
\usepackage{algpseudocode}
\usepackage{booktabs}
\usepackage{multirow}
\usepackage[table]{xcolor}
\usepackage{graphicx}
\usepackage{float}
\usepackage{dblfloatfix}

\begin{document}

\title{When 1+1 Backfires: Rethinking Relevance-Diversity Balancing for Sample-Aware Visual Token Reduction}

\author{Yihang Huang}
\affiliation{%
  \institution{State Key Laboratory of AI Safety, Institute of Computing Technology, Chinese Academy of Sciences}
  \institution{School of Computer Science and Technology, University of Chinese Academy of Sciences}
  \city{Beijing}
  \country{China}
}
\orcid{0009-0003-2986-7484}
\email{huangyihang25s@ict.ac.cn}

\author{Peisong Wen}
\affiliation{%
  \institution{School of Computer Science and Technology, University of Chinese Academy of Sciences}
  \city{Beijing}
  \country{China}
}
\orcid{0000-0002-8881-3609}
\email{wenpeisong@ucas.ac.cn}

\author{Chenhao Zhou}
\affiliation{%
  \institution{State Key Laboratory of AI Safety, Institute of Computing Technology, Chinese Academy of Sciences}
  \institution{School of Computer Science and Technology, University of Chinese Academy of Sciences}
  \city{Beijing}
  \country{China}
}
\orcid{0009-0007-8189-7911}
\email{zhouchenhao25@mails.ucas.ac.cn}

\author{Difu Feng}
\affiliation{%
  \institution{State Key Laboratory of AI Safety, Institute of Computing Technology, Chinese Academy of Sciences}
  \institution{School of Computer Science and Technology, University of Chinese Academy of Sciences}
  \city{Beijing}
  \country{China}
}
\orcid{0009-0000-9855-4535}
\email{fengdifu24@mails.ucas.ac.cn}

\author{Qianqian Xu}
\correspondingauthor
\affiliation{%
  \institution{State Key Laboratory of AI Safety, Institute of Computing Technology, Chinese Academy of Sciences}
  \institution{Beijing Academy of Artificial Intelligence}
  \city{Beijing}
  \country{China}
}
\orcid{0000-0002-3512-7277}
\email{xuqianqian@ict.ac.cn}

\renewcommand{\shortauthors}{Yihang Huang, Peisong Wen, Chenhao Zhou, Difu Feng, and Qianqian Xu}
\begin{abstract}
  Visual token compression has emerged as a critical technique for reducing the inference cost of Vision-Language Models. Existing methods select tokens based on text relevance, visual diversity, or a fixed combination of both---treating the relevance-diversity balance as a universal constant independent of the input. We conduct a systematic analysis to challenge this assumption. At the dataset level, grid-searching the optimal relevance-diversity weight per benchmark reveals substantial cross-task variation. More critically, we further maximize per-sample answer likelihood to recover sample-level oracle weights, finding that their distributions not only differ markedly across benchmarks but also exhibit high within-benchmark variance. Based on these observations, we propose SAVER, a framework for \textbf{S}ample-\textbf{A}ware relevance-diversity balanced \textbf{V}isual tok\textbf{E}n \textbf{R}eduction. We first derive a Geometric Intent Prior (GIP) from the spatial concentration of text-relevant tokens to infer a sample-specific relevance-diversity weight, and then apply this weight to jointly select tokens via a Weighted Text Centroid (WTC) for relevance scoring and a Visual Marginal Coverage Gain (VMCG) for diversity. SAVER requires no additional training and achieves sample-aware balancing entirely at inference time.
  Extensive experiments demonstrate that SAVER consistently outperforms state-of-the-art baselines across diverse benchmarks, retaining 96.5\% of full-model performance on LLaVA-1.5-7B with only 11.1\% of the original visual tokens, and further generalizing to LLaVA-NeXT and Mini-Gemini.
\end{abstract}




\maketitle

\section{Introduction}

Vision-Language Models (VLMs)~\cite{llava1.5,qwenvl,qwen2vl,minigemini,llavanext,flamingo,internvl,deepseekvl,blip2,hua2024reconboost,COMBINER}, which jointly process visual and textual inputs, have achieved significant progress across a wide range of multimodal tasks, including visual question answering, image captioning, and document understanding. To faithfully encode image semantics, these models represent visual inputs as a large number of tokens. The resulting long token sequences directly inflate the input length, incurring $O(N^2)$ attention complexity in the Transformer~\cite{transformer} backbone and creating a critical computational bottleneck that limits the practical deployment of LVLMs.

To mitigate this bottleneck, visual token reduction~\cite{fastv,pdrop,saint,mustdrop,sparsevlm,dart,prunesid,holov,cdpruner,mmtok,llavaprumerge,VTW,visa} has emerged as a promising direction. Existing methods can be broadly grouped into three families. Text-relevance methods~\cite{fastv,pdrop} prioritize visual tokens that align strongly with the prompt, but often over-focus on a few high-response regions and discard peripheral context. Visual-diversity methods~\cite{saint,divprune} instead maximize coverage of the image content, but remain agnostic to the query and may waste token budget on irrelevant regions. More recent methods~\cite{mustdrop,sparsevlm,dart} attempt to integrate both objectives within a unified pipeline. However, they do not explicitly model the sample-dependent trade-off between relevance preservation and visual coverage. As observed in prior work~\cite{mob}, such input-agnostic balancing does not consistently outperform single-objective baselines, a phenomenon termed ``$1+1 < 1$.'' These findings suggest that the central challenge lies not merely in combining relevance and diversity, but in estimating an appropriate per-sample allocation between the two objectives.

In this work, we investigate this phenomenon through a systematic two-level analysis (Section~\ref{sec:motivation}). Specifically, we formulate the token selection score as a weighted combination of text relevance and visual diversity, controlled by a scalar weight $\theta$, and greedily select tokens in descending order of this score. At the dataset level, we find that each benchmark exhibits a clear optimal relevance-diversity weight that delivers noticeably better performance than alternative allocations, while the locations of these optima vary substantially across tasks. We further investigate the source of this variation at the instance level. Since answer correctness is discrete and does not provide a continuous optimization signal, we use per-sample answer likelihood as a surrogate objective to recover oracle weights for individual inputs. The recovered oracle weights further show that the sample-level preference is consistent with the relevance-diversity tendency observed at the dataset level. \textbf{These results suggest that effective token reduction fundamentally requires sample-aware relevance-diversity balancing.}

Closing this gap is non-trivial. A practical sample-aware token reduction method must satisfy two key requirements. \textbf{1).} The prior must be derived from existing knowledge and the current input, without relying on any task-specific or dataset-level context unavailable at inference time. This is precisely the limitation of MoB~\cite{mob}: although it provides a principled theoretical account of the relevance-diversity trade-off, its allocation strategy must be configured per dataset. In open-world deployment, however, the dataset identity of an incoming input is simply unknown. \textbf{2).} The prior must be obtained within the early prefill stage from shallow semantics alone, without auxiliary modules or additional inference beyond the compression pipeline. Relying on an extra predictor, or waiting for more complete semantics from deeper layers, would introduce substantial pre-pruning cost and compromise the efficiency gains of token reduction, thereby undermining its very purpose as an inference acceleration technique.

Crucially, we observe that the spatial distribution of high-relevan-ce visual tokens provides an effective cue for the desired relevance-diversity balance. When high-relevance tokens are spatially concentrated, the textual semantics are strongly aligned with a specific local region of the image, which is likely to contain the information most relevant to answering the query. When high-relevance tokens are broadly dispersed, the model cannot localize truly relevant visual tokens with high confidence, making relevance scores less reliable as a basis for selection. In this case, a larger diversity weight is preferred to preserve broad visual coverage and maintain prediction accuracy. 

Driven by these observations, we propose a Sample-Aware Rele-vance-Diversity Balanced Token Reduction framework. Its core component is the Geometric Intent Prior (GIP), a training-free prior that quantifies the text-visual coupling of each sample from the spatial covariance of its text-relevant visual tokens. GIP produces a continuous scalar that adaptively modulates the relative weight of each objective, enabling sample-specific objective balancing at negligible cost. Complementing GIP, we introduce two objective-specific improvements: a norm-weighted text centroid for text relevance estimation, free of cross-modal attention and compatible with FlashAttention~\cite{flashattention}; and a facility location formulation for visual coverage whose marginal gain jointly accounts for novelty and representativeness.
Extensive evaluations demonstrate that our method retains 96.5\% of full-model performance using only 11\% of the visual tokens, consistently outperforming existing approaches under equivalent compression budgets.

Our main contributions are summarized as follows:
\begin{itemize}
    \item \textbf{Analysis for Relevance-Diversity Balancing.} We conduct a systematic two-level empirical analysis showing that the optimal relevance-diversity balance varies substantially across both tasks and individual samples, establishing that per-sample balancing of the relevance-diversity objectives is necessary for effective visual token compression.
    \item \textbf{Solution to Sample-Aware Objective Balancing.} We propose GIP, a training-free, efficient solution to sample-aware relevance-diversity balancing, which derives a sample-specific weight from the spatial distribution of text-relevant tokens.
    \item \textbf{SAVER Framework.} We propose a Weighted Text Centroid for text relevance scoring and a Visual Marginal Coverage Gain for diversity-aware token selection. By dynamically balancing these two objectives via GIP, our proposed framework SAVER achieves superior performance over competitive baselines across multiple benchmarks and generalizes to diverse model architectures.
\end{itemize}

\begin{figure*}[t]
    \centering
    \includegraphics[width=\linewidth]{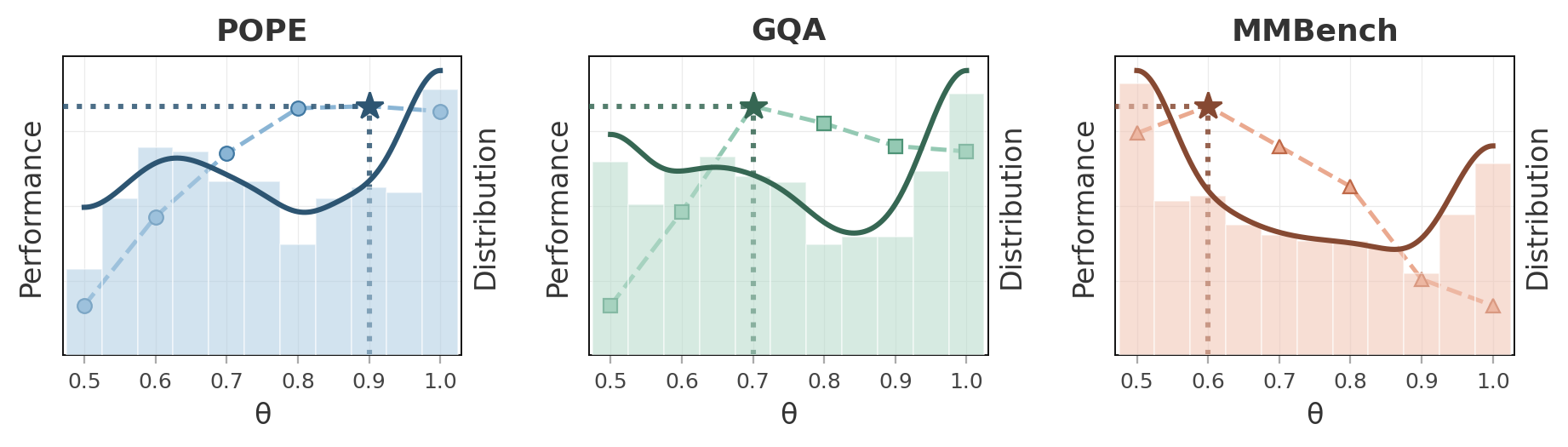}
    \caption{Dataset-level performance vs.\ $\theta$ (dashed line) and instance-level oracle weight distributions $\theta_i^*$ (solid curve and histogram) for POPE, GQA, and MMBench. POPE favors high $\theta$ (diversity-dominant); MMBench favors low $\theta$ (relevance-dominant); GQA peaks at intermediate values. Per-sample distributions closely correspond to dataset-level optima.}
    \label{fig:motivation}
\end{figure*}

\section{Related Work}

\subsection{Large Vision-Language Models}

Large Vision-Language Models (LVLMs)~\cite{llava1.5,qwenvl,qwen2vl,minigemini,llavanext} have emerged from the foundational success of Large Language Models (LLMs)~\cite{llama,gpt4,qwen,vicuna,mixtral}. The standard LVLM pipeline consists of three core components. A visual encoder, typically a pre-trained vision transformer such as CLIP~\cite{clip}, extracts a sequence of visual tokens from the input image. A text tokenizer independently processes the natural language instruction. Both modalities are then projected into a shared embedding space and fed into a core LLM for joint multimodal reasoning. This architecture has achieved strong performance across a wide range of tasks, including visual question answering, image captioning, and document understanding. Despite these advances, faithfully encoding image semantics requires generating a large number of visual tokens. This directly inflates the input sequence length, which in turn substantially increases computational overhead and bottlenecks inference efficiency.

\subsection{Token Reduction Methods}

Existing token reduction methods largely fall into two categories: text-relevance and visual coverage. Text-relevance methods use cross-attention maps to select tokens that align with the text prompt. FastV filters tokens by raw attention scores~\cite{fastv}, and PyramidDrop extends this into a progressive, layer-wise pruning scheme~\cite{pdrop}. These methods quickly locate core semantic regions. Yet they tend to over-select tokens near these hubs and discard important peripheral context. Visual coverage methods instead maximize the semantic diversity of the selected token set. SAINT prunes via bipartite graph node degrees~\cite{saint}, and DivPrune formulates redundancy reduction as a Max-Min Diversity Problem~\cite{divprune}. These methods recover the global visual context more faithfully. However, they are task-agnostic. They do not consider the text instruction during selection, which can waste token budget on irrelevant regions.

Recent work has attempted to combine both objectives within a unified pipeline~\cite{sparsevlm,dart}. Most such approaches use fixed weighting schemes that do not adapt to the input. In practice, the optimal balance between relevance and diversity varies considerably across different questions. Li et al.\ formalize this trade-off using $\epsilon$-covering theory~\cite{mob}. They show that multimodal datasets split into two regimes---weak and strong text-visual coupling---and that each regime requires different allocation parameters for best performance. A limitation of their method is that it requires knowing the target dataset in advance to set these parameters. This limits applicability in open-ended settings where the task intent is unknown at inference time.

\section{Motivation}
\label{sec:motivation}

This section presents a systematic empirical study of the relevance-diversity trade-off in visual token compression. We first formulate the dataset-level optimal relevance-diversity weight and identify how it varies across benchmarks. To gain deeper insight into the origins of this cross-dataset variation, we further investigate the distribution of instance-level optimal weights within each benchmark. Since no continuous metric directly reflects token selection quality at the sample level, we introduce per-sample answer likelihood as a surrogate objective to recover these instance-level weights. Based on both levels of analysis, we report empirical observations that motivate the design of our method.

\noindent\textbf{Dataset-Level Optimal Weight.}
Given a multimodal input with visual token set $V = \{v_1,\ldots,v_N\}$ and text token sequence $T$, we define the per-token selection score as a weighted combination of text relevance $\mathrm{Rel}(\cdot,T)$ and visual diversity $\mathrm{Div}(\cdot)$:
\begin{equation}
  \mathrm{Score}_\theta(v_j,\, T) = (1-\theta) \cdot \mathrm{Rel}(v_j,\, T) \;+\; \theta \cdot \mathrm{Div}(v_j),
  \label{eq:score}
\end{equation}
where $\theta \in [0,1]$ controls the relevance-diversity balance, with larger $\theta$ corresponding to greater diversity weight. Under a fixed retention budget $K$, we select the top-$K$ tokens by score:
\begin{equation}
  V_\theta = \mathrm{Top}_K\bigl\{\mathrm{Score}_\theta(v_j,\,T)\bigr\}_{j=1}^{N}.
  \label{eq:select}
\end{equation}
For each benchmark $\mathcal{D}_k$ with task evaluation metric $\mathcal{M}$, the dataset-level optimal weight is defined as:
\begin{equation}
  \theta_k^* = \arg\max_{\theta \in [0,1]}\; \frac{1}{|\mathcal{D}_k|} \sum_{(x,\,Y)\in \mathcal{D}_k} \mathcal{M}\!\left(f_\phi\!\left(V_\theta,\, T\right),\, Y\right),
  \label{eq:dataset_opt}
\end{equation}
where $Y$ is the ground-truth answer. 

\noindent\textbf{Instance-Level Oracle Weight.}
To probe the distribution of optimal weights at the sample level, we define the \emph{oracle weight} for each sample $i$ as the $\theta$ that maximizes the likelihood of the correct answer under the LLM $\Phi$. Formally, let $\mathcal{L}_i(\theta)$ denote the negative log-likelihood (NLL) of the ground-truth answer given $\mathcal{V}_\theta$ and $\mathcal{T}$:
\begin{equation}
  \mathcal{L}_i(\theta) = -\frac{1}{|y|}\sum_{t=1}^{|y|} \log P_{f_\phi}\!\left(y_t \mid V_\theta,\, T,\, y_{<t}\right).
  \label{eq:nll}
\end{equation}
The instance-level oracle weight is then defined as:
\begin{equation}
  \theta_i^* = \arg\min_{\theta \in [0,1]}\; \mathcal{L}_i(\theta).
  \label{eq:oracle}
\end{equation}
We recover $\theta_i^*$ for sampled instances from each benchmark $\mathcal{D}_k$ and analyze the resulting distributions.

\begin{figure*}[t]
    \centering
    \includegraphics[width=\linewidth]{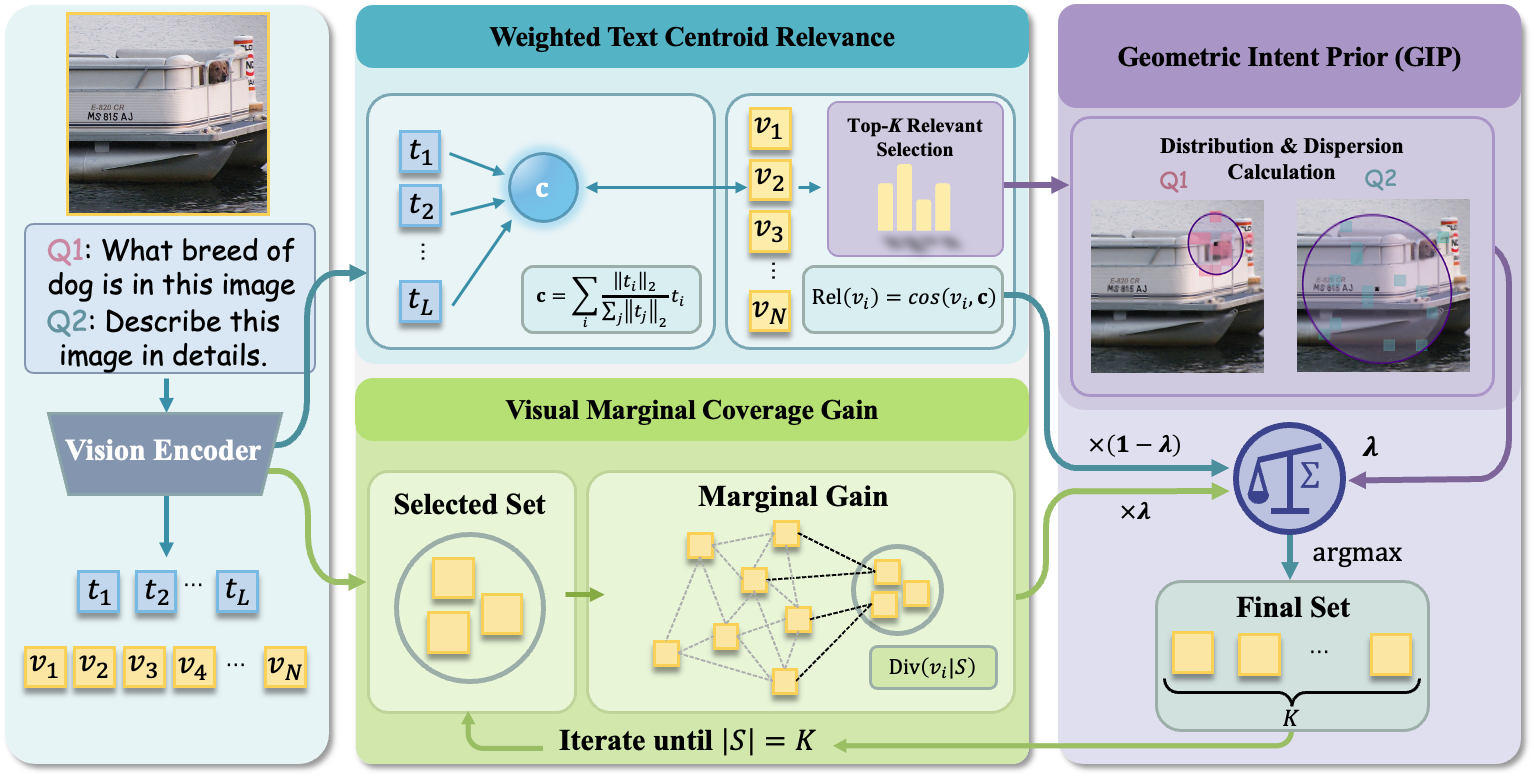}
    \caption{Overview of the SAVER framework. Given a multimodal input, we compute the Geometric Intent Prior (GIP) from the spatial distribution of high-relevance visual tokens, which dynamically weights text relevance (via WTC) and visual diversity (via VMCG) in a unified greedy selection process.}
    \label{fig:framework}
\end{figure*}

\noindent\textbf{Empirical Observations.}
As illustrated by the dashed lines in Figure~\ref{fig:motivation}, $\theta_k^*$ varies substantially across benchmarks: POPE~\cite{pope} favors high $\theta$ (diversity-dominant), MMBench~\cite{mmbench} favors low $\theta$ (relevance-dominant), and GQA~\cite{gqa} peaks at an intermediate value. The per-sample distributions $\{\theta_i^*\}$  are well aligned with their respective dataset-level optima, and this pattern is consistent with the intuitive characteristics of each task. POPE primarily probes object hallucination by querying small or absent objects, requiring the model to survey diverse regions across the visual field; MMBench queries fine-grained local details, where focused retrieval of the relevant region is decisive. Critically, both distributions also exhibit substantial within-benchmark variance, indicating that the optimal balance is not a property of the task alone but varies at the level of individual inputs. 

\algnewcommand\algorithmicHyperparameters{\textbf{Hyperparameters:}}
\algnewcommand\Hyperparameters{\item[\algorithmicHyperparameters]}
\begin{algorithm}[t]
    \caption{Sample-Aware Token Reduction}
    \label{alg:main}
    \begin{algorithmic}[1]
    \Require $V=\{v_i\}_{i=1}^N$, $T=\{t_i\}_{i=1}^L$, budget $K$
    \Hyperparameters $\gamma,\; k,\; b$
    \Ensure $S$ with $|S|=K$
    \State $\mathbf{c} \leftarrow \sum_i \frac{\|t_i\|_2}{\sum_j\|t_j\|_2}\, t_i$;\quad $\mathrm{Rel}(v_i) \leftarrow \cos(v_i, \mathbf{c})\ \forall\,i$
    \State $\mathcal{G} \leftarrow \det\!\left(\operatorname{Cov}\!\left(\{(x_i,y_i): v_i\in\operatorname{TopK}(V,\lfloor 0.02N\rfloor)\}\right)\right)$
    \State $\tilde{\mathcal{G}} \leftarrow \mathcal{N}_{\mathcal{G}}(\mathcal{G})$
    \State $k_{\mathrm{GIP}} \leftarrow k\,\mathcal{N}_{\mathcal{D}}\!\left(1-\frac{1}{N(N-1)}\sum_{p\neq q}\cos(v_p,v_q)\right)$
    \State $\theta \leftarrow \operatorname{sigmoid}\!\left(k_{\mathrm{GIP}} \cdot \tilde{\mathcal{G}} + b\right)$
    \State $S \leftarrow \emptyset$
    \While{$|S| < K$}
        \State $\mathrm{Div}(v_i{\mid}S) \leftarrow {-}\gamma\max_{s\in S}K_{si}$
        \Statex \hspace{3.2em} $+ (1{-}\gamma)\displaystyle\sum_{j\notin S,\,j\neq i}\max\!\bigl(0,\; K_{ij}{-}\max_{s\in S}K_{sj}\bigr),\quad\forall\,v_i\notin S$
        \State $v^* \leftarrow \arg\max_{v_i \notin S}\; (1{-}\theta)\,\mathrm{Rel}(v_i) + \theta\,\mathrm{Div}(v_i{\mid}S)$
        \State $S \leftarrow S \cup \{v^*\}$
    \EndWhile
    \State \Return $S$
    \end{algorithmic}
\end{algorithm}

\section{Methodology}

In this section, we present our Sample-Aware Unified Token Reduction framework. We first introduce a Geometric Intent Prior (GIP) in Section~\ref{sec:gip}, which infers the sample-specific text-visual coupling from the spatial distribution of text-relevant visual tokens and adaptively balances the two selection objectives. We then describe our reformulated text relevance objective via a Weighted Text Centroid (WTC) in Section~\ref{sec:wtc}, and our visual diversity objective via a Visual Marginal Coverage Gain (VMCG) in Section~\ref{sec:vmcg}. Finally, we present the unified token selection procedure in Section~\ref{sec:unified}. The overall framework is illustrated in Figure~\ref{fig:framework}.

\begin{table*}[!t]
  \centering
  \caption{Comparison with state-of-the-art token reduction methods on LLaVA-1.5-7B across three compression budgets.}
  \label{tab:main}
  \resizebox{\textwidth}{!}{%
  \begin{tabular}{lcccccccc}
  \toprule
  Method & GQA & MMBench & MME & POPE & ScienceQA & TextVQA & OCRBench & Avg. \\
  \midrule
  \rowcolor{gray!15}\multicolumn{9}{c}{\textit{No compression (576 tokens)}} \\
  LLaVA-1.5-7B & 61.9 & 64.3 & 1862 & 85.9 & 69.5 & 58.2 & 297 & 100\% \\
  \midrule
  \rowcolor{gray!15}\multicolumn{9}{c}{\textit{192 tokens (66.7\% reduction)}} \\
  FastV (ECCV'24)~\cite{fastv}         & 52.7 & 61.2 & 1612 & 64.8 & 67.3 & 52.5 & 291 & 89.6\% \\
  SparseVLM (ICML'25)~\cite{sparsevlm} & 57.6 & 62.5 & 1721 & 83.6 & 69.1 & 56.1 & 292 & 96.3\% \\
  PyramidDrop (CVPR'25)~\cite{pdrop}   & 57.1 & 63.2 & 1766 & 82.3 & 68.8 & 56.1 & 290 & 96.3\% \\
  DivPrune (CVPR'25)~\cite{divprune}   & 59.9 & 62.4 & 1752 & 87.0 & 68.9 & 57.0 & 303 & 98.3\% \\
  DART (EMNLP'25)~\cite{dart}          & 60.0 & 63.6 & 1856 & 82.8 & 69.8 & 57.4 & 296 & 98.7\% \\
  VisionZip (CVPR'25)~\cite{visionzip} & 59.3 & 63.0 & 1783 & 85.3 & 68.9 & 57.3 & --  & 97.7\% \\
  PruneSID (ICLR'26)~\cite{prunesid}   & 60.1 & 63.7 & 1791 & 86.9 & 68.5 & 56.7 & 294 & 98.4\% \\
  SAVER (Ours)                         & 61.2 & 63.8 & 1855 & 85.7 & 69.3 & 57.8 & 300 & \textbf{99.6\%} \\
  \midrule
  \rowcolor{gray!15}\multicolumn{9}{c}{\textit{128 tokens (77.8\% reduction)}} \\
  FastV (ECCV'24)~\cite{fastv}         & 49.6 & 56.1 & 1490 & 59.6 & 60.2 & 50.6 & 285 & 83.8\% \\
  SparseVLM (ICML'25)~\cite{sparsevlm} & 56.0 & 60.0 & 1696 & 80.5 & 67.1 & 54.9 & 280 & 93.4\% \\
  PyramidDrop (CVPR'25)~\cite{pdrop}   & 56.0 & 61.1 & 1644 & 82.3 & 68.3 & 55.1 & 287 & 94.2\% \\
  DivPrune (CVPR'25)~\cite{divprune}   & 59.2 & 62.3 & 1752 & 86.6 & 68.0 & 54.5 & 283 & 96.3\% \\
  DART (EMNLP'25)~\cite{dart}          & 58.7 & 63.2 & 1840 & 80.1 & 69.1 & 56.4 & 296 & 97.3\% \\
  VisionZip (CVPR'25)~\cite{visionzip} & 57.6 & 62.0 & 1762 & 83.2 & 68.9 & 56.8 & --  & 96.3\% \\
  PruneSID (ICLR'26)~\cite{prunesid}   & 58.8 & 62.1 & 1749 & 86.5 & 68.3 & 54.7 & 275 & 95.9\% \\
  SAVER (Ours)                         & 60.9 & 63.6 & 1811 & 85.5 & 69.2 & 57.2 & 287 & \textbf{98.4\%} \\
  \midrule
  \rowcolor{gray!15}\multicolumn{9}{c}{\textit{64 tokens (88.9\% reduction)}} \\
  FastV (ECCV'24)~\cite{fastv}         & 46.1 & 48.0 & 1256 & 48.0 & 51.1 & 47.8 & 245 & 72.9\% \\
  SparseVLM (ICML'25)~\cite{sparsevlm} & 52.7 & 56.2 & 1505 & 75.1 & 62.2 & 51.8 & 180 & 82.8\% \\
  PyramidDrop (CVPR'25)~\cite{pdrop}   & 41.9 & 33.3 & 1092 & 55.9 & 68.6 & 45.9 & 250 & 72.1\% \\
  DivPrune (CVPR'25)~\cite{divprune}   & 57.6 & 59.5 & 1638 & 85.8 & 68.6 & 52.9 & 275 & 93.7\% \\
  DART (EMNLP'25)~\cite{dart}          & 55.9 & 60.6 & 1765 & 73.9 & 69.8 & 54.4 & 270 & 92.9\% \\
  VisionZip (CVPR'25)~\cite{visionzip} & 55.1 & 60.1 & 1690 & 77.0 & 69.0 & 55.5 & --  & 92.9\% \\
  PruneSID (ICLR'26)~\cite{prunesid}   & 57.1 & 58.8 & 1733 & 83.8 & 67.8 & 54.2 & 273 & 93.8\% \\
  SAVER (Ours)                         & 60.0 & 61.8 & 1758 & 83.8 & 68.9 & 56.2 & 282 & \textbf{96.5\%} \\
  \bottomrule
  \end{tabular}%
  }
\end{table*}

\subsection{Preliminaries}

Formally, we denote the $N$ visual tokens produced by the visual encoder as $V = \{v_i\}_{i=1}^{N}$ and the $L$ text tokens from the text tokenizer as $T = \{t_i\}_{i=1}^{L}$. The concatenated sequence $[V; T]$ is fed into the LLM decoder $f_\phi$, which autoregressively generates the response $y = f_\phi([V; T])$. Visual token reduction selects a subset $S \subseteq V$ with $|S| = K \ll N$ at an early layer of the LLM, so that the decoder operates on the compressed sequence $[S; T]$ instead, reducing prefilling cost.

\subsection{Geometric Intent Prior}\label{sec:gip}

As established in Section~\ref{sec:motivation}, the optimal relevance-diversity balance is inherently sample-dependent, yet existing methods treat it as a fixed, input-agnostic hyperparameter. Accessing deep cross-modal representations from the LLM backbone could in principle provide rich cues for estimating the sample-specific balance, but doing so requires a full forward pass—the very computation that token reduction is designed to avoid. A practical solution must therefore rely on a proxy derivable from shallow representations in the early prefill stage, before the full depth of the LLM is engaged.

We propose to use the spatial distribution of high-relevance visual tokens as such a proxy. The key observation is that the spatial spread of these tokens reflects the degree to which the input query can be resolved by a localized region of the image. When high-relevance tokens are spatially concentrated, the query is grounded in a specific local region, and a higher relevance weight is appropriate to retrieve detail-oriented tokens from that region. When they are broadly dispersed, the text tokens lack strong semantic alignment with any specific visual region, making relevance scores unreliable as a selection criterion; in this case, a higher diversity weight is warranted to ensure broad spatial coverage and allow the model to attend to global visual details. This spatial geometry is fully determined from shallow token representations within the LLM, requiring no additional training or full forward pass. We formalize this as the Geometric Intent Prior (GIP).

Formally, let $\mathcal{V} = \{v_i\}_{i=1}^{N}$ denote the set of visual tokens and let $\operatorname{Rel}(v_i)$ denote the text relevance score of each token (defined in Section~\ref{sec:wtc}). We select the top-$k$ most text-relevant tokens to form the anchor set:
\begin{equation}
    \mathcal{V}_k = \bigl\{v_i \in \mathcal{V} \mid \operatorname{Rel}(v_i) \in \operatorname{TopK}(\{\operatorname{Rel}(v_j)\}_{j=1}^{N},\, k)\bigr\}.
\end{equation}
We set $k = \lfloor 0.02 N \rfloor$, i.e., the top 2\% of visual tokens by relevance score. As validated in our experiments, tokens in this percentile exhibit significantly higher relevance scores than the remainder and carry rich semantic content, making them reliable indicators of the query's visual grounding. 

Each token $v_i \in \mathcal{V}_k$ corresponds to a spatial patch at 2D coordinate $\mathbf{p}_i = (x_i, y_i) \in \mathbb{R}^2$. Let $\mathcal{P} = \{\mathbf{p}_i\}_{i=1}^{k}$ denote the corresponding spatial point set. We compute the $2 \times 2$ spatial covariance matrix:
\begin{equation}
    \boldsymbol{\Sigma} = \frac{1}{k} \sum_{i=1}^{k} (\mathbf{p}_i - \bar{\mathbf{p}})(\mathbf{p}_i - \bar{\mathbf{p}})^\top,
\end{equation}
where $\bar{\mathbf{p}} = \frac{1}{k}\sum_{i=1}^{k} \mathbf{p}_i$ is the spatial centroid of $\mathcal{P}$. We define the Geometric Intent Prior as its determinant:
\begin{equation}
    \mathcal{G} = \det(\boldsymbol{\Sigma}).
\end{equation}
$\mathcal{G}$ equals the \emph{generalized variance} of $\mathcal{P}$, measuring the area of the covariance ellipsoid spanned by the anchor tokens. A small $\mathcal{G}$ indicates that the high-relevance tokens are spatially concentrated; a large $\mathcal{G}$ indicates broad spatial dispersion. How $\mathcal{G}$ is mapped to the objective weighting scalar $\theta$ is described in Section~\ref{sec:unified}.

\subsection{Text Relevance via Weighted Text Centroid}\label{sec:wtc}

Previous token reduction methods~\cite{fastv,sparsevlm,dart} estimate text relevance from cross-modal attention maps within the LLM, measuring the relevance of each visual token by its accumulated attention weight toward text tokens. However, two limitations hinder this approach in practice. First, modern inference pipelines increasingly adopt FlashAttention~\cite{flashattention}, which fuses the attention kernel and does not materialize the full attention matrix, rendering attention-based relevance estimation inapplicable. Second, LLM attention maps are susceptible to positional biases~\cite{sparsevlm}, where early tokens attract disproportionately high attention irrespective of their semantic content, introducing systematic noise into the relevance scores.

We instead propose to estimate text relevance via the semantic centroid of the text token embeddings, which is both attention-free and compatible with any inference backend. A naive unweighted mean, however, is susceptible to interference from function words and punctuation tokens, which contribute equally despite carrying little semantic content. Empirically, content words exhibit consistently higher $L_2$ norms in the embedding space than function words and punctuation. We exploit this property to define a norm-weighted centroid, termed Weighted Text Centroid (WTC), which naturally suppresses semantically weak tokens.

Formally, let $\mathcal{T} = \{t_i\}_{i=1}^{L}$ denote the text token embeddings. The normalized $L_2$ norm of each token serves as its weight:
\begin{equation}
    w_i = \frac{\|t_i\|_2}{\sum_{j=1}^{L} \|t_j\|_2}.
\end{equation}
The Weighted Text Centroid is defined as:
\begin{equation}
    \mathbf{c} = \sum_{i=1}^{L} w_i\, t_i.
\end{equation}
The text relevance score of each visual token $v_i$ is then its cosine similarity with $\mathbf{c}$:
\begin{equation}
    \operatorname{Rel}(v_i) = \frac{v_i \cdot \mathbf{c}}{\|v_i\|\, \|\mathbf{c}\|}.
\end{equation}
This formulation requires only a single inner product per visual token and introduces no dependency on attention scores or LLM intermediate states.

\begin{table*}[t]
  \centering
  \caption{Generalization results on LLaVA-NeXT.}
  \label{tab:generalization}
  \resizebox{0.92\textwidth}{!}{%
  \begin{tabular}{lcccccccc}
  \toprule
  Method & \#Tokens & GQA & MMBench & POPE & MME & ScienceQA & SEEDBench & Avg. \\
  \midrule
  LLaVA-NeXT (Full)                  & 2880 & 64.2 & 67.9 & 86.4 & 1842 & 70.2 & 70.2 & 100\% \\
  \midrule
  VisionZip (CVPR'25)~\cite{visionzip} & 640  & 61.3 & 66.3 & 86.3 & 1787 & 68.1 & 66.7 & 97.0\% \\
  PruneSID (ICLR'26)~\cite{prunesid}   & 640  & 61.6 & 64.2 & 86.3 & 1795 & 68.3 & 67.3 & 96.8\% \\
  SAVER (Ours)                         & 640 & 63.6 & 65.8 & 86.4 & 1802 & 69.8 & 68.5 & \textbf{98.5\%} \\
  \midrule
  VisionZip (CVPR'25)~\cite{visionzip} & 320  & 59.3 & 63.1 & 82.1 & 1702 & 67.3 & 63.4 & 93.2\% \\
  PruneSID (ICLR'26)~\cite{prunesid}   & 320  & 60.5 & 63.0 & 83.1 & 1754 & 67.3 & 65.0 & 94.5\% \\
  SAVER (Ours)                         & 320 & 62.5 & 65.0 & 85.0 & 1706 & 69.4 & 66.2 & \textbf{96.2\%} \\
\bottomrule
  \end{tabular}%
  }
\end{table*}

\begin{table*}[t]
  \centering
  \caption{Generalization results on Mini-Gemini.}
  \label{tab:gen_qwen}
  \resizebox{0.92\textwidth}{!}{%
  \begin{tabular}{lcccccccc}
  \toprule
  Method & \#Tokens & GQA & MMBench & POPE & MME & ScienceQA & SEEDBench & Avg. \\
  \midrule
  Mini-Gemini (Full)                   & 576 & 62.4 & 69.3 & 85.8 & 1841 & 70.7 & 69.7 & 100\% \\
  \midrule
  VisionZip (CVPR'25)~\cite{visionzip} & 128 & 58.7 & 68.1 & 78.5 & 1841 & 70.0 & 65.6 & 96.2\% \\
  PruneSID (ICLR'26)~\cite{prunesid}   & 128 & 60.1 & 66.6 & 82.4 & 1821 & 70.7 & 66.5 & 97.1\% \\
  SAVER (Ours)                         & 128 & 60.6 & 67.2 & 84.1 & 1831 & 71.0 & 67.2 & \textbf{98.1\%} \\
  \midrule
  VisionZip (CVPR'25)~\cite{visionzip} & 64  & 55.8 & 65.9 & 69.6 & 1737 & 70.7 & 61.7 & 91.4\% \\
  PruneSID (ICLR'26)~\cite{prunesid}   & 64  & 58.3 & 63.1 & 76.0 & 1735 & 70.6 & 63.6 & 93.1\% \\
  SAVER (Ours)                         & 64 & 58.7 & 64.8 & 81.0 & 1742 & 70.0 & 64.4 & \textbf{94.7\%} \\
  \bottomrule
  \end{tabular}%
  }
\end{table*}

\subsection{Visual Diversity via Visual Marginal Coverage Gain}\label{sec:vmcg}

Visual diversity in token selection has been approached in two main ways, each with a distinct failure mode. Max-min diversity methods~\cite{divprune} prioritize tokens that are maximally distant from the current selection, which makes them sensitive to outliers: a single token lying in a visually uninformative region can dominate the selection simply by being far from others. Alternatively, coverage-based methods favor tokens that best represent dense groups of similar visual patches, which tends to over-allocate budget to large homogeneous regions (e.g., background) and under-represent compact foreground structures rich in semantic detail.

Inspired by the facility location problem in combinatorial optimization, we propose a selection criterion that avoids both failure modes. In the facility location analogy, each unselected token is a ``client'' that must be served by the nearest selected token (``facility''). Adding a new token is only beneficial if it meaningfully improves coverage for clients that are currently underserved. This naturally penalizes both redundant selections and outliers.

Formally, let $\mathbf{K} \in \mathbb{R}^{N \times N}$ denote the cosine similarity matrix of the visual tokens, where each entry is defined as:
\begin{equation}
    K_{ij} = \frac{v_i \cdot v_j}{\|v_i\|\,\|v_j\|}.
\end{equation}
Given the already-selected set $S$, the Visual Marginal Coverage Gain (VMCG) of a candidate token $v_i$ is:
\begin{equation}
\begin{aligned}
    \mathrm{Div}(v_i \mid S) &= \gamma \cdot \max\!\left(0,\, K_{ii} - \max_{s \in S} K_{si}\right) \\
    &\quad + (1{-}\gamma)\cdot\!\!\sum_{v_j \notin S,\,j \neq i}\!\max\!\left(0,\, K_{ij} - \max_{s \in S} K_{sj}\right).
\end{aligned}
\end{equation}
Since $K_{ii}=1$ for cosine-normalized features, and dropping the additive constant $\gamma$, this reduces to:
\begin{equation}
    \mathrm{Div}(v_i \mid S) = \underbrace{-\gamma \max_{s \in S} K_{si}}_{\text{novelty}} + \underbrace{(1{-}\gamma)\!\!\sum_{v_j \notin S,\,j \neq i}\!\max\!\left(0,\, K_{ij} - \max_{s \in S} K_{sj}\right)}_{\text{representativeness}},
\end{equation}
where $\gamma \in (0,1)$ balances novelty against representativeness. The novelty term measures how dissimilar $v_i$ is from the closest token already in $S$. In the representativeness term, the $\max(0, \cdot)$ operator encodes a key principle from facility location: an unselected token $v_j$ is only ``served'' by $v_i$ if $v_i$ is more similar to $v_j$ than any token currently in $S$—if a better-matched facility already exists, $v_j$ contributes nothing to the score of $v_i$. Consequently, $\mathrm{Div}(v_i \mid S)$ is high only when $v_i$ is both novel with respect to the current selection and genuinely representative of a large portion of tokens that remain poorly covered, naturally suppressing both redundant and outlier selections.

\subsection{Sample-Aware Token Selection}\label{sec:unified}

Given $\mathrm{Rel}(v_i)$ and $\mathrm{Div}(v_i \mid S)$ defined above, the unified selection score for each candidate token is:
\begin{equation}
    \mathrm{Score}(v_i \mid S) = (1 - \theta)\cdot \mathrm{Rel}(v_i) + \theta \cdot \mathrm{Div}(v_i \mid S),
\end{equation}

Let $V=\{v_i\}_{i=1}^{N}$ be the visual tokens of image $I$. For a statistic $X$, define $\mathcal{N}_{X}(x)=\operatorname{clip}((x-X_{\min})/(X_{\max}-X_{\min}))$; hence $\tilde{\mathcal{G}}=\mathcal{N}_{\mathcal{G}}(\mathcal{G})$.
\begin{samepage}
Raw image diversity is defined as
\begin{equation}
    \mathcal{D}(I)=1-\frac{1}{N(N-1)}\sum_{i\neq j}\frac{v_i^\top v_j}{\lVert v_i\rVert_2\,\lVert v_j\rVert_2}.
\end{equation}
\end{samepage}
Whereas GIP captures prompt-conditioned spatial variation among relevant tokens, $\mathcal{D}(I)$ measures intrinsic visual variation within the image. We normalize it as $\mathrm{DIV}(I)=\mathcal{N}_{\mathcal{D}}(\mathcal{D}(I))$ and use it only to scale the GIP mapping:
\begin{equation}
    k_{\mathrm{GIP}}=k\cdot\mathrm{DIV}(I).
\end{equation}
The sample-specific weight is then
\begin{equation}
    \theta=\operatorname{sigmoid}\!\left(k_{\mathrm{GIP}}\cdot\tilde{\mathcal{G}}+b\right).
\end{equation}
The bounds $\mathcal{G}_{\min},\mathcal{G}_{\max},\mathcal{D}_{\min},\mathcal{D}_{\max}$ used by $\mathcal{N}_{\mathcal{G}}$ and $\mathcal{N}_{\mathcal{D}}$ are estimated from a small held-out calibration subset; clipping prevents out-of-range samples from producing abnormal weights. Here, $k$ and $b$ are global linear coefficients in the sigmoid pre-activation that scale and shift $\tilde{\mathcal{G}}$ into an appropriate range before mapping to $\theta \in (0,1)$.

At each greedy step, the token with the highest score is appended to $S$, and the diversity scores are updated incrementally. Selection proceeds until $|S| = K$; the full procedure is given in Algorithm~\ref{alg:main}.

\section{Experiments}

\subsection{Experimental Setup}

\noindent\textbf{Benchmarks.}
We evaluate on a diverse set of image-based multimodal benchmarks covering general visual understanding: GQA~\cite{gqa}, MMBench~\cite{mmbench}, MME~\cite{mme}, POPE~\cite{pope}, ScienceQA~\cite{sqa}, SEEDBen-ch~\cite{seedbench} and TextVQA~\cite{textvqa}. We additionally include OCRBench~\cite{ocrbench} to assess OCR and document understanding. All benchmarks are evaluated under their standard protocols.

\noindent\textbf{Baselines.}
We compare SAVER against representative token reduction methods spanning single-objective and multi-objective design strategies. Text-relevance-based methods include FastV~\cite{fastv} and PyramidDrop~\cite{pdrop}. Visual-diversity-based methods include DivPrune~\cite{divprune}, VisionZip~\cite{visionzip}, and PruneSID~\cite{prunesid}. Methods combining both objectives include SparseVLM~\cite{sparsevlm} and DART~\cite{dart}.

\noindent\textbf{Model architectures.}
We validate SAVER primarily on LLaVA-1.5-7B~\cite{llava1.5} and further evaluate on LLaVA-NeXT-7B~\cite{llavanext} and Mini-Gemini~\cite{minigemini} to assess generalizability. All models are evaluated following their original inference configurations, without any additional training.

\subsection{Main Results}

\noindent\textbf{LLaVA-1.5-7B.}
LLaVA-1.5-7B is a widely adopted open-source multimodal baseline that resizes input images to a fixed resolution and encodes them into a fixed set of 576 visual tokens via a CLIP vision encoder. We evaluate at three retention budgets $K \in \{64, 128, 192\}$. As shown in Table~\ref{tab:main}, SAVER ranks first under all three settings. At $K=64$, SAVER achieves 96.5\%, surpassing PruneSID and VisionZip by 2.7 and 3.6 points. 
At $K=128$, SAVER attains 98.4\%, maintaining a consistent lead over all baselines. At $K=192$, SAVER retains 99.6\% of full-model performance with negligible degradation.

\noindent\textbf{LLaVA-NeXT.}
LLaVA-NeXT encodes both the full image and local crops at high resolution, yielding up to 2880 visual tokens and substantially stronger fine-grained perception than single-scale models. We evaluate at $K \in \{320, 640\}$ and report results in Table~\ref{tab:generalization}. SAVER achieves 98.5\% at 640 tokens and 96.2\% at 320 tokens, outperforming PruneSID (96.8\%/94.5\%) and VisionZip (97.0\%/93.2\%).

\noindent\textbf{Mini-Gemini.}
Mini-Gemini employs a dual-encoder architecture that simultaneously extracts high- and low-resolution visual features and fuses them into 576 tokens, enabling richer multi-scale representations. We evaluate at $K \in \{64, 128\}$ and report results in Table~\ref{tab:gen_qwen}. SAVER achieves 98.1\% at 128 tokens and 94.7\% at 64 tokens, consistently ahead of PruneSID (97.1\%/93.1\%) and VisionZip (96.2\%/91.4\%).

\subsection{Efficiency Analysis}

We compare inference efficiency against SparseVLM~\cite{sparsevlm} and DAR-T~\cite{dart}, both of which, like SAVER, apply token reduction during the LLM prefilling stage. We measure peak GPU memory, inference throughput, and POPE accuracy on LLaVA-1.5-7B at $K=64$ retained tokens. As reported in Table~\ref{tab:efficiency}, SAVER matches DART in both peak memory and throughput, while achieving a POPE score of 83.8 compared to DART's 73.9. SparseVLM, despite consuming more GPU memory than the uncompressed model, attains only 75.1 on POPE. These results demonstrate that SAVER's accuracy gains introduce no additional computational overhead.

\begin{table}[t]
\centering
\caption{Efficiency comparison on LLaVA-1.5-7B.}
\label{tab:efficiency}
\resizebox{0.95\columnwidth}{!}{%
\begin{tabular}{lcccc}
\toprule
Method & \#Tok. & Peak GPU$\downarrow$ & Throughput$\uparrow$ & POPE$\uparrow$ \\
\midrule
LLaVA-1.5-7B (Full)        & 576 & 14.69 & 11.21 & 85.9 \\
\midrule
SparseVLM~\cite{sparsevlm} & 64  & 18.84 & 12.09 & 75.1 \\
DART~\cite{dart}            & 64  & 14.33 & 17.17 & 73.9 \\
SAVER (Ours)                & 64  & 14.33 & 17.11 & 83.8 \\
\bottomrule
\end{tabular}%
}
\end{table}

\begin{table}[t]
    \centering
    \caption{Ablation study on LLaVA-1.5-7B.}
    \label{tab:ablation}
    \begin{tabular}{lcccc}
    \toprule
    Method & POPE & GQA & MMBench & Avg. \\
    \midrule
    WTC relevance               & 73.7 & 56.3 & 61.9 & 91.0\% \\
    VMCG diversity               & 83.4 & 58.7 & 59.9 & 95.0\% \\
    WTC + VMCG(static)            & 84.0 & 59.5 & 60.4 & 95.9\% \\
    SAVER (Ours)                  & 83.8 & 60.0 & 61.8 & \textbf{96.9\%} \\
    \bottomrule
    \end{tabular}
\end{table}

\begin{figure}[t]
  \centering
    \includegraphics[width=0.9\linewidth]{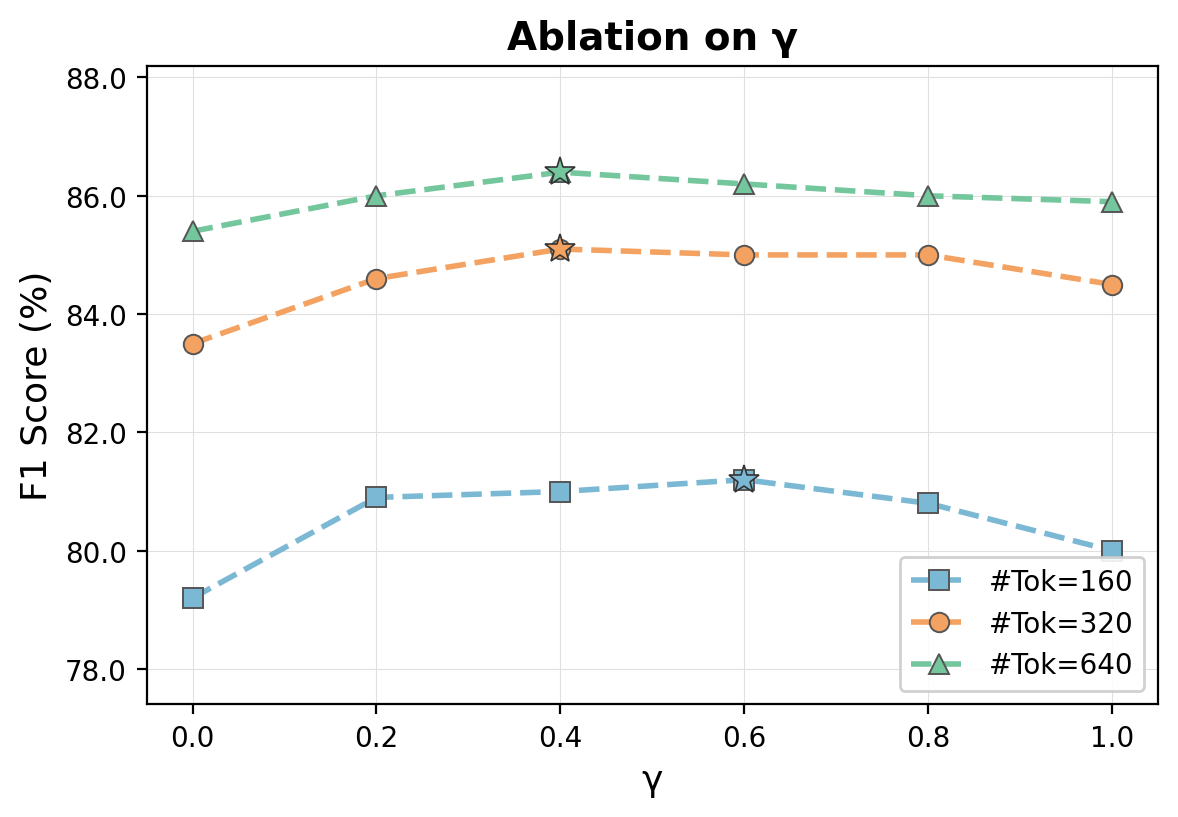}
    \caption{Ablation on $\gamma$ across different token budgets on LLaVA-NeXT-7B (POPE F1). The effect is amplified under higher compression ratios.}
    \label{fig:gamma}
\end{figure}

\subsection{Ablation Study}

\noindent\textbf{Component ablation.}
We ablate the core design choices of SAVER on LLaVA-1.5-7B at $K=64$ tokens retained, as shown in Table~\ref{tab:ablation}.
Using relevance alone yields the lowest average (91.0\%): without diversity, the selected tokens cluster around query-relevant regions and fail to cover the broader visual context, which is reflected in a notably low POPE score of 73.7.
Diversity alone improves the average to 95.0\% and raises POPE to 83.4 through wider spatial coverage, but the absence of text guidance leads to a drop in MMBench (59.9), where responses depend on precise localization of query-relevant details.
The static combination improves the average to 95.9\%, but remains below the sample-aware variant because a fixed relevance-diversity weight cannot accommodate the varying demands of different inputs.
SAVER achieves the best average of 96.9\% by dynamically inferring a sample-specific balance via GIP, confirming that adaptive, instance-level balancing is the key factor behind its consistent gains.

\noindent\textbf{Discussion on the Novelty--Representativeness Trade-off.}
The novelty-representativeness trade-off coefficient $\gamma$ requires more care. As shown in Figure~\ref{fig:gamma}, overly small $\gamma$ values under-emphasize novelty and bias selection toward redundant regions, while overly large $\gamma$ values over-prioritize outlier tokens at the expense of global coverage. Moderate values yield the best performance in practice. The optimal $\gamma$ may also vary with the token retention budget.

\section{Conclusion}

We present a systematic empirical analysis of the relevance--diversity trade-off in visual token reduction, demonstrating that the optimal objective balance varies substantially across both tasks and individual inputs and cannot be captured by any fixed strategy. Motivated by this finding, we propose SAVER, a training-free framework centered on the Geometric Intent Prior (GIP), which infers a sample-specific relevance-diversity weight directly from the spatial distribution of high-relevance tokens. GIP is paired with a Weighted Text Centroid for attention-free relevance estimation and a Visual Marginal Coverage Gain for diversity-aware selection. Extensive experiments show that SAVER consistently outperforms competitive baselines across all compression budgets, while matching their inference efficiency, confirming that instance-level adaptive balancing is a principled approach to visual token reduction.

\begin{acks}
  This work was supported in part by National Natural Science Foundation of China: 62525212 and U23B2051,  in part by Youth Innovation Promotion Association CAS,  in part by the Strategic Priority Research Program of the Chinese Academy of Sciences, Grant No. XDB0680201, in part by the China National Postdoctoral Program for Innovative Talents under Grant BX20250377, in part by the China Postdoctoral Science Foundation under Grant 2026M791710, in part by the Beijing Major Science and Technology Project under Contract Z251100008125059, and in part by Beijing Academy of Artificial Intelligence (BAAI).
\end{acks}

\bibliographystyle{ACM-Reference-Format}
\bibliography{sample-base}

\appendix

\end{document}